\documentclass[letterpaper,onecolumn]{aamas2013_extendedabstract}
\usepackage[T1]{fontenc}
\usepackage[latin9]{inputenc}
\usepackage{amstext}
\usepackage{graphicx}
\usepackage{esint}

\makeatletter

\pdfpageheight\paperheight
\pdfpagewidth\paperwidth

\usepackage{algpseudocode}
\usepackage{xcolor}

\usepackage{times}

\newcommand{\superscript}[1]{\ensuremath{^{\textrm{#1}}}}

\makeatother

\begin{document}

\title{Multi-agent RRT{*}: Sampling-based Cooperative Pathfinding}

\author{\numberofauthors{1} 
\alignauthor Michal {\v C}{\' a}p\superscript{1}\thanks{The research was supported by the ARTEMIS Joint Undertaking under the number 269336-2 and by the Czech Republic Ministry of Education, Youth and Sports, grant no. 7H11102 (D3CoS).}, Peter Nov{\'a}k\superscript{2}, Ji\v{r}\'{i} Vok\v{r}\'{i}nek\superscript{1} and Michal  P\v{e}chou{\v{c}}ek\superscript{1} \\ 
~\\ 
\superscript{1}\affaddr{Agent Technology Center, DCSE, FEE, Czech Technical University in Prague, Czech Republic}  \\
\superscript{2}\affaddr{Algorithmics, EEMCS, Delft University of Technology, The Netherlands} }
\maketitle
\begin{abstract}
Cooperative pathfinding is a problem of finding a set of non-con\-flic\-ting
trajectories for a number of mobile agents. Its applications include
planning for teams of mobile robots, such as autonomous aircrafts,
cars, or underwater vehicles. The state-of-the-art algorithms for
cooperative pathfinding typically rely on some heuristic forward-search
pathfinding technique, where A{*} is often the algorithm of choice.
Here, we propose MA-RRT{*}, a novel algorithm for multi-agent path
planning that builds upon a recently proposed asymptotically-optimal
sampling-based algorithm for finding single-agent shortest path called
RRT{*}. We experimentally evaluate the performance of the algorithm
and show that the sam\-pling-based approach offers better scalability
than the classical for\-ward-search approach in relatively large,
but sparse environments, which are typical in real-world applications
such as multi-aircraft collision avoidance.
\end{abstract}



\section{Introduction}

The problem of collision avoidance for mobile robots, such as aircrafts
can be modeled as an instance of \emph{cooperative pathfinding}, a
relatively well studied problem of finding a set of non-conflicting
trajectories for a number of mobile agents. 

The straightforward, but complete approach to the problem is to search
the solution in what we call a joint-state space. The state space
is constructed as the Cartesian product of the state spaces of the
individual agents. It is typically searched using some heuristic forward
search algorithm, such as A{*}. The performance of forward search
algorithms hinges on low branching factor of the search space, which
is in joint-state spaces often exponential in the number of agents.
Suppose for example that an agent can move in four directions and
that the problem involves six agents. Then, there is $4^{6}=4096$
possible joint actions at each timestep! Clearly, the completeness
of such an approach is traded for a prohibitive computational cost. 

Recently, Karaman and Frazzoli~\cite{RRTStar:Karaman.Frazzoli:IJRR11}
introduced a novel any-time sampling-based motion planning algorithm
that offers good scalability to high-dimensional environments, while
at the same time it guarantees convergence to an optimal solution.
In this paper, we introduce MA-RRT{*}, a novel sampling-based algorithm
for cooperative pathfinding, the main contribution of this paper.
The algorithm searches for the plan of agents' movements in their
joint-state space, but replaces the A{*}-based heuristic search in
the joint-state space by RRT{*}. We extensively evaluate the performance
and solution quality produced by the algorithm and show that for sparsely
populated large environments the sampling algorithm outperforms Standley
and Korf's optimal anytime algorithm (OA)~\cite{StandleyK11} in
terms of runtime and success rate, while still maintaining reasonable
quality of the solution.

\section{Problem Formulation}

To allow fair comparison with the OA algorithm, which is defined only
for agents moving on graphs, we use the following definition of a
cooperative pathfinding problem. Consider $n$ agents operating in
an Euclidean space. The motion model of the agent $i$ is described
by a corresponding motion graph denoted as $G_{i}^{M}=(W_{i},M_{i})$.
The starting positions of all agents are given as an $n$-tuple $(s_{1},\ldots,s_{n})$,
where $s_{i}\in W_{i}$ is the starting waypoint of agent $i$. Similarly,
$(d_{1},\ldots,d_{n})$ is an $n$-tuple of destination waypoints
$d_{i}\in W_{i}$ of each agent. The task is to find a sequence of
motion primitives, i.e., a path $p_{i}$ in the motion graph $G_{i}^{M}$
for each agent $i$, such that $\mathit{start}(p_{i})=s_{i}$ and
$\mathit{end}(p_{i})=d_{i}$ and the paths are separated, that is,
$\forall j,k,t:\, j\neq k\Rightarrow dist(p_{i}[t],p_{j}[t])>d_{\mathit{sep}}$,
with $d_{\mathit{sep}}$ being the required separation distance. As
the solution quality metric we use the sum of times each of the agents
spends outside his destination waypoint.

\section{The Algorithm}

The RRT{*} algorithm is designed for continuous state spaces in which
it can efficiently find a path from a given start state to a given
target region by incrementally building a tree that is rooted at the
start state and spans towards randomly sampled states from some given
state space. Once the tree first reaches the goal region, the algorithm
can follow its edges backwards to obtain the first feasible path from
the start state to the target region. However, even after the first
solution is returned, the algorithm does not stop, but instead continues
extending the tree by drawing new random samples, which leads to incremental
discovery of new lower-cost paths.

We use identical approach to find the shortest path in a motion graph.
The core difference is that in continuous version, two samples can
be connected if they are mutually visible. In the graph version of
RRT{*}, two samples can be connected if it is possible to find a valid
path between them by heuristic-guided greedy search in the motion
graph.

The multi-agent version of graph RRT{*} (MA-RRT{*}) is identical to
the graph version of RRT{*}, except that it searches for a shortest
path in a graph that represents the joint-state space of all agents.
The returned solution is then a collision-free joint plan containing
a path for each agent. 

The performance of the algorithm on an average problem instance can
be improved by biasing the sampling distribution to favor regions
around optimal paths of the individual agents that are more likely
to contain high quality solutions. We call this variant an informed-sampling
MA-RRT{*}.

\section{Evaluation}

We compared the performance of the unbiased version of MA-RRT{*} (MA-RRT{*})
and informed-sampling MA-RRT{*} (isMA-RRT{*}) with A{*} search in
joint-state space (JA) and optimal anytime algorithm (OA) in terms
of scalability and solution quality. All three algorithms were implemented
in \emph{Java} in a common framework. 

We evaluated the performance of the algorithms on the following set
of synthetic problem instances. The agents move on a square-shaped
grid-like motion graph, where the waypoints were placed on the grid
having the step of 1~meter and the motion primitives were straight
moves at the constant speed of 1~m/s connecting the vertices in the
4-neighborhood. Furthermore, a 1~second long waiting motion primitive
was available at each waypoint. We randomly removed 10~percent of
the vertices of the motion graph to represent obstacles. A unique
start waypoint and unique destination waypoint was chosen randomly
for each agent. Finally, for each such instance we checked whether
all agents can reach their destinations to ensure that the instance
admits a solution. 

The set of tested problem instances contained instances that varied
in the size of the grid and in the number of agents. We used the following
values of the two parameters. Grid sizes: 10x10, 30x30, 50x50, 70x70,
90x90. Numbers of agents: 1, 2, 3, 4, 5, 6, 7, 8, 9, 10. The separation
distance was set to a constant 0.8. The problem instance set contained
120 random instances (with random obstacles and random start and destination
positions) for each combination of the grid size and the number of
agents. Thus, in total, the experiment included 6000 different problem
instances. Each of the algorithms was executed on every instance with
the runtime limit of 5 seconds. The experiments were performed on
\emph{HotSpot 1.6 64-bit Java VM} running on \emph{AMD FX-8150} 3.6
GHz CPU.

\subsection{Results}

\begin{figure}[h]
\begin{centering}
\includegraphics[bb=0bp 12bp 410bp 200bp,clip,scale=0.55]{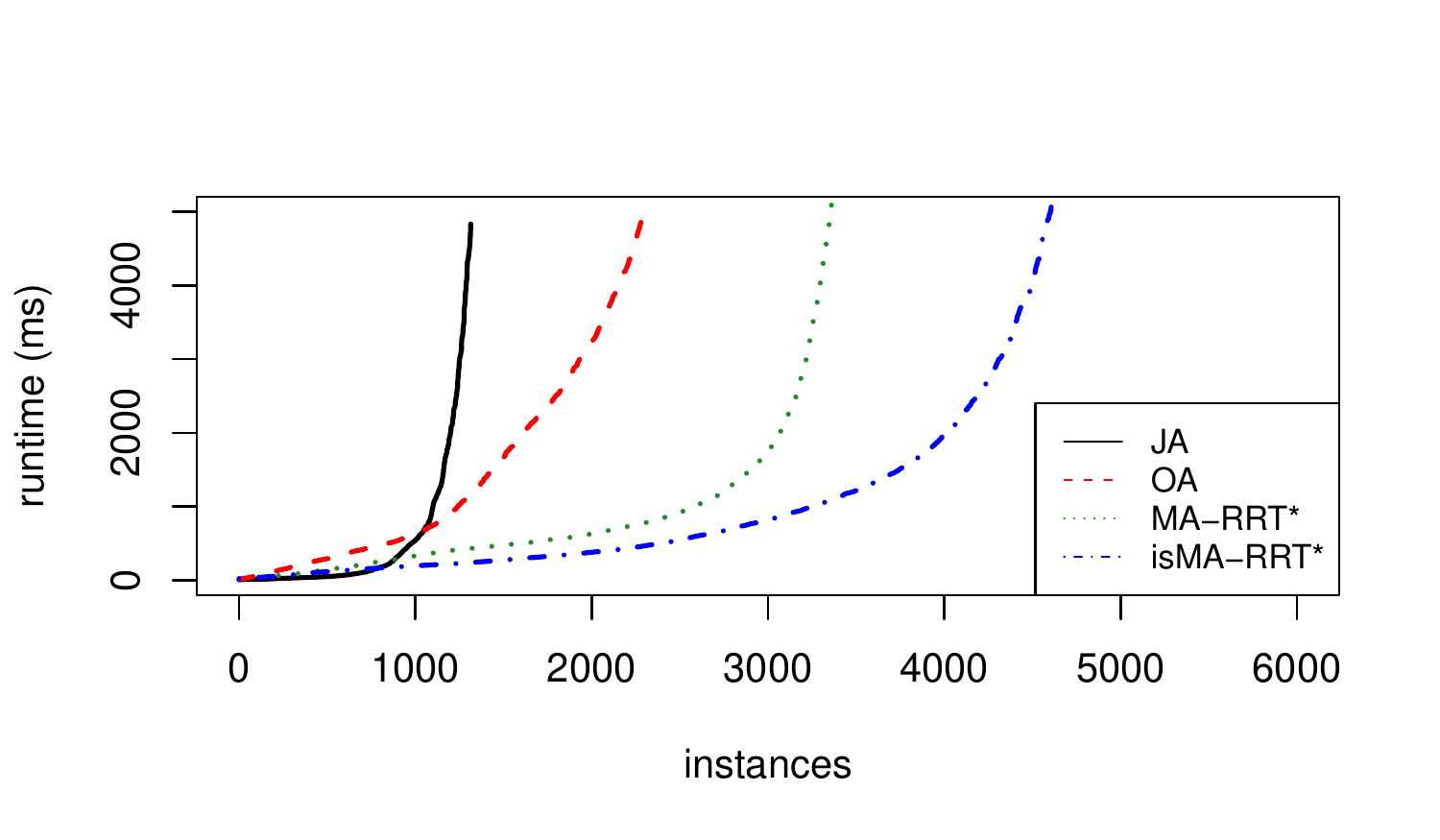}
\par\end{centering}

\caption{\label{fig:Performance-curve}First-solution performance curve}
\end{figure}

To convey how well the algorithms performed on the evaluation set
of problem instances, we plot the performance curves (proposed in~\cite{StandleyK11})
for each algorithm. We recorded the runtime to find the first valid
solution to the problem instance for each algorithm. Then, we sorted
the instances according to the runtime for each algorithm independently.
The results are plotted in Figure~\ref{fig:Performance-curve}. On
the x-axis is the index of instance in the algorithm's sorted sequence,
on the y-axis is the runtime the algorithm needed to find the first
solution to that problem instance. It should be noted that the ordering
of the instances is different for each algorithm. The x-position of
the last point in the performance curve can be interpreted as the
number of instances of the total 6000 instances the algorithm solved
in the runtime limit of 5 seconds. We can see that JA resolved 21\%
of the instances, OA 38\%, MA-RRT{*} 56\% and isMA-RRT{*} 77 \% of
instances from our problem instance set. 

\begin{figure}[h]
\begin{centering}
\includegraphics[bb=0bp 45bp 410bp 200bp,clip,scale=0.55]{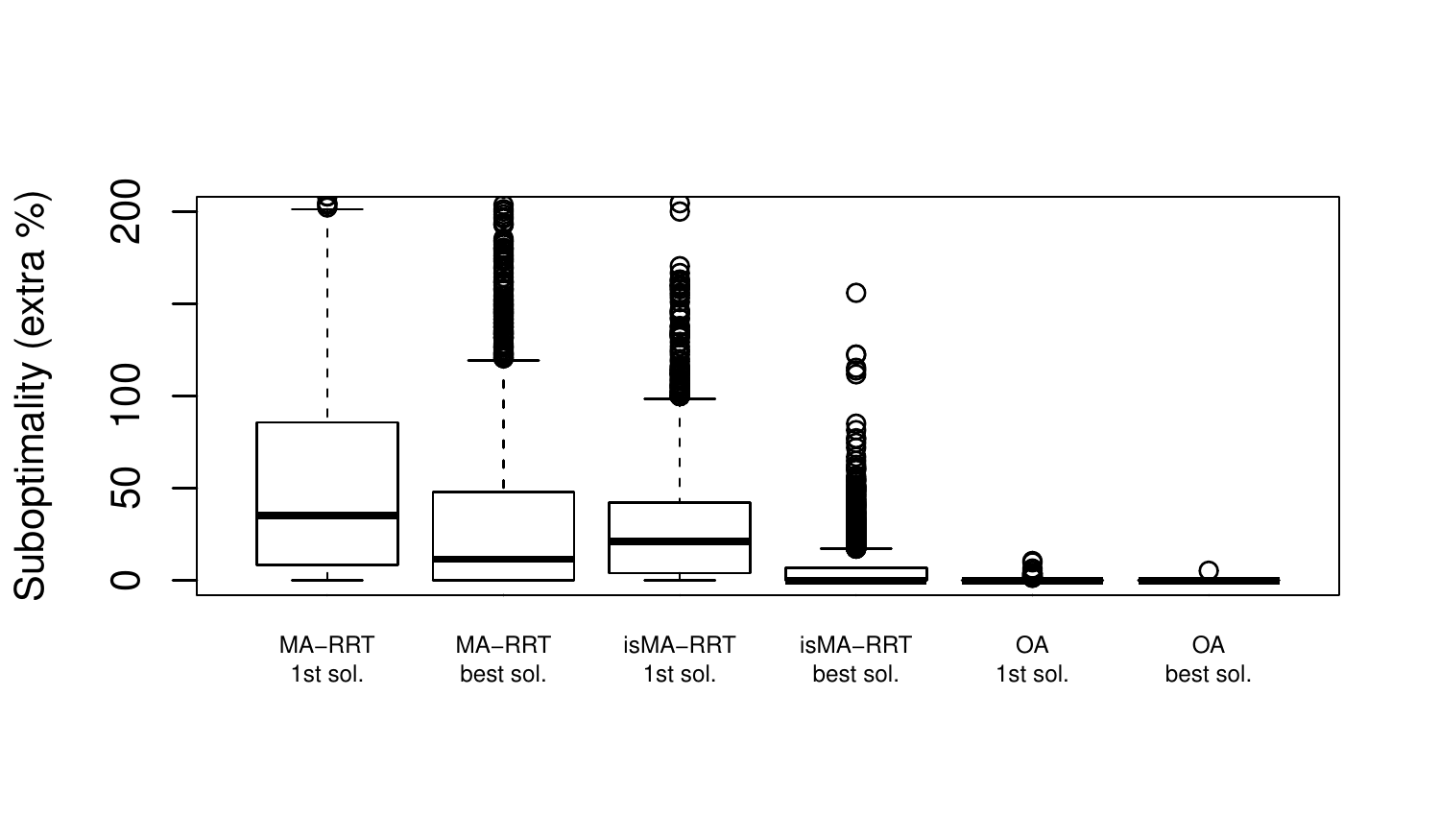}
\par\end{centering}

\caption{\label{fig:Solution-quality}Solution quality}
\end{figure}

Figure~\ref{fig:Solution-quality} shows the comparison of relative
solution quality for the anytime algorithms, JA is not plotted since
it always returns optimal solutions. For all algorithms we show the
quality of the first returned solution and the quality of the best
solution found within the 5~seconds runtime limit. The suboptimality
is measured only on a subset of instances for which either JA or OA
returned provably optimal solution (in our case 2438 instances). The
suboptimality measure is expressed in percentage points as follows:

\[
\textrm{suboptimality}=\left(\frac{\textrm{cost\ of\,\ returned\,\ solution}}{\textrm{cost\,\ of\,\ optimal\,\ solution}}-1\right)\cdot100.
\]

\section{Conclusion}

In this paper we proposed MA-RRT{*}, an anytime algorithm for solving
cooperative pathfinding problems. Our experiments demonstrate the
limits of the forward-search based approaches to cooperative pathfinding
in large, but sparse environments. Our results show that these instances
can be efficiently solved using one of our sampling-based algorithms
for the price of a slight decrease in the solution quality. 

\bibliographystyle{plain}
\bibliography{bib}

\begin{thebibliography}{1}

\bibitem{RRTStar:Karaman.Frazzoli:IJRR11}
Karaman and Frazzoli.
\newblock Sampling-based algorithms for optimal motion planning.
\newblock {\em International Journal of Robotics Research}, 30(7):846--894,
  June 2011.

\bibitem{StandleyK11}
Trevor~Scott Standley and Richard~E. Korf.
\newblock Complete algorithms for cooperative pathfinding problems.
\newblock In Toby Walsh, editor, {\em IJCAI}, pages 668--673. IJCAI/AAAI, 2011.

\end{thebibliography}

\end{document}